# Controllable Diffusion-Based Lesion Inpainting for Scalable Histopathology Data Augmentation

Mohamad Koohi-Moghadam[1*], Mohammad-Ali Nikouei Mahani[1], Rex K.H. Au-Yeung[2], Raymond Yu O[2], Monalyn Marabi[3], Piyapharom Intarawichian[3], Fabian Z. X. Lean[4], Andrew Ferguson[5], Kyongtae Tyler Bae[1*]

[1] Department of Diagnostic Radiology, Li Ka Shing Faculty of Medicine, The University of Hong Kong, Pok Fu Lam, Hong Kong SAR, PR China
[2] Department of Pathology, Li Ka Shing Faculty of Medicine, The University of Hong Kong, Hong Kong SAR, China
[3] Department of Anatomical and Cellular Pathology, Prince of Wales Hospital, The Chinese University of Hong Kong, Hong Kong SAR, China.
[4] Department of Infectious Diseases and Public Health, Jockey Club College of Veterinary Medicine and Life Sciences, City University of Hong Kong, Hong Kong SAR, China
[5] CityU Veterinary Diagnostic Laboratory Co., Ltd., City University of Hong Kong, Hong Kong SAR, China

Corresponding author: koohi@hku.hk, baekt@hku.hk

## Abstract

Expert-annotated training data remains the critical bottleneck for AI in histopathology, particularly for rare pathologies where even dozens of cases may be unavailable. While data augmentation offers a solution, existing methods fail to generate sufficiently realistic lesion morphologies that preserve tissue-specific architectures. Here we present PathoGen, a diffusion-based generative model enabling controllable, high-fidelity lesion inpainting into benign histopathology images. We validate PathoGen across four datasets representing kidney, skin, breast, and prostate pathology. Quantitative assessment confirms PathoGen outperforms state-of-the-art baselines in image fidelity and distributional similarity. Evaluation by six expert pathologists revealed that synthetic images by PathoGen were only marginally distinguished from real tissue image slightly above chance (57.75% accuracy), demonstrating strong perceptual realism of PathoGen-generated lesions. PathoGen achieved the highest win rate (35.4%) when pathologists ranked generation quality against all baselines. Crucially, augmenting training sets with PathoGen-synthesized lesions improves segmentation Dice scores by up to 0.18 compared to traditional augmentations, with maximum benefit in data-scarce regimes. By simultaneously generating realistic morphology and pixel-level annotations, PathoGen effectively addresses both data scarcity and annotation cost—two critical bottlenecks in computational pathology development.

## Introduction

The integration of artificial intelligence into histopathology has demonstrated remarkable potential for improving diagnostic accuracy, reducing pathologist workload, and enabling quantitative tissue analysis at unprecedented scales [1, 2]. Deep learning models have achieved expert-level performance in detecting malignancies [3], predicting treatment responses [4], and discovering novel prognostic biomarkers [5]. However, the development of robust, generalizable AI systems for histopathology remains fundamentally constrained by a critical challenge: the scarcity of expertly annotated training data, particularly for rare pathologies, early-stage lesions, and diagnostically challenging cases [6, 7]. Unlike natural image datasets where millions of labeled examples are readily available, histopathology suffers from inherent data limitations. Pixel-level annotation of tissue lesions requires extensive domain expertise, with pathologists spending hours delineating complex morphological structures at cellular resolution [8]. This annotation burden is compounded by patient privacy concerns, institutional data silos, and the rarity of certain disease presentations—some pathological subtypes may have only dozens of documented cases worldwide [9]. Consequently, AI models trained on limited datasets exhibit poor generalization to new institutions, staining protocols, and patient populations, creating a significant barrier to clinical deployment [10].

Data augmentation has emerged as a cornerstone strategy for mitigating data scarcity in medical imaging [11]. Traditional geometric transformations—rotation, flipping, scaling, and color jittering—have shown modest improvements in model robustness [12]. However, these approaches merely recombine existing visual information without introducing novel pathological variations. They cannot generate new lesion morphologies, synthesize rare disease presentations, or create examples that bridge gaps in the training distribution. This limitation is particularly acute in histopathology, where diagnostic features span multiple spatial scales from subcellular organelles to tissue architecture, and where subtle morphological variations carry critical clinical significance [13, 14]. Recent advances in generative modeling offer a promising avenue for synthetic data generation. Generative Adversarial Networks (GANs) have been applied to histopathology image synthesis [15, 16], demonstrating the feasibility of creating visually plausible tissue images. However, GAN-based approaches face substantial challenges in this domain. Training instability often results in mode collapse, where generators produce limited diversity in lesion morphologies [17], which can manifest as artifacts at tissue boundaries and inconsistent staining patterns. Additionally, existing generative methods typically lack fine-grained control over lesion-specific attributes such as placement, size, and morphology—limiting their utility for targeted data augmentation in specific diagnostic scenarios [18].

Diffusion models have recently emerged as a powerful class of generative models that address many limitations of GANs [19]. By learning to iteratively denoise data through a reverse diffusion process, these models achieve superior sample quality, training stability, and mode coverage [20]. Diffusion models have demonstrated exceptional performance in natural image synthesis [21], medical image reconstruction [22], and various image-to-image translation tasks [23]. Their iterative refinement process is particularly well-suited for generating fine-grained details—a critical requirement for histopathology where diagnostic decisions depend on subcellular features. Moreover, recent work on conditional diffusion models has shown that spatial control can be achieved through various conditioning mechanisms [24-26], suggesting their potential for controllable lesion synthesis. Despite these promising characteristics, diffusion models remain largely unexplored for histopathology data augmentation. Existing applications have focused primarily on reconstruction tasks [27] or unconditional whole-image generation [28], rather than the targeted, controllable lesion inpainting required for effective augmentation. The challenge lies in developing conditioning strategies that enable precise spatial control over lesion placement while ensuring that generated lesions exhibit natural integration with surrounding benign tissue, maintain realistic cellular morphology, and preserve tissue-specific architectural patterns.

Here we introduce PathoGen, a diffusion-based generative model specifically designed for controllable lesion inpainting in histopathology images. PathoGen adapts a pretrained latent diffusion backbone to the histopathology domain through a self-supervised inpainting objective. Given a benign tissue image and a spatial mask indicating the desired lesion location, PathoGen synthesizes a realistic lesion within the specified region while maintaining seamless integration with surrounding tissue. We demonstrate PathoGen's effectiveness across four diverse histopathology datasets representing clinically important diagnostic challenges: kidney (glomerular disease), melanoma, breast tumour (subtyping) and prostate tumour (cancer grading). These datasets span different tissue types and diagnostic tasks, providing a rigorous test of generalization. To assess the perceptual realism of the synthesized lesions, we conducted a reader study in which multiple expert pathologists evaluated both the distinguishability of synthetic from real tissue and the relative quality of PathoGen against competing generative methods. When PathoGen-generated synthetic images are used to augment training data for lesion segmentation models, we observe consistent and substantial improvements in segmentation accuracy across all tissue types, with the most pronounced gains in low-data regimes where manual annotation is most resource-intensive. Notably, PathoGen automatically provides pixel-level annotations for all generated lesions, simultaneously addressing both data scarcity and annotation cost—two critical bottlenecks in computational pathology development.

## Materials and Methods

### Datasets

We trained and evaluated PathoGen across four diverse histopathology datasets representing distinct tissue types, disease morphology, and diagnostic challenges. The TIGER dataset [29] comprises digital pathology whole-slide images (WSIs) of HER2-positive and triple-negative breast cancer (TNBC) with expert annotations of tissue compartments, lymphocytes, and plasma cells; for our experiments the target structure was the invasive tumor region. The KPI dataset [30] addresses chronic kidney disease (CKD) through Periodic acid–Schiff (PAS)–stained WSIs containing annotated glomeruli, which served as the target structure. The PUMA dataset [31] includes regions of interest from both primary and metastatic melanoma scanned at high magnification, with the melanoma tumor as the target structure. The RINGS dataset [32] focuses on prostate gland segmentation in histopathological images, encompassing both healthy and pathological structures with varying degrees of glandular degeneration, with the glandular structures as the target. With the exception of the PAS-stained KPI dataset, all datasets were hematoxylin and eosin (H&E) stained **(Supplementary Table 1)**. For each dataset we reserved 20% of WSIs as a held-out test set to evaluate model performance on completely unseen cases, with the remaining 80% used for training. This split was performed at the slide level so that test images originate from different patients and slides than those used during training, providing a rigorous assessment of generalization. From the training slides we extracted 1024×1024-pixel non-overlapping patches scanned at a resolution of 226 nm/pixel, producing 10,000 training patches per dataset (40,000 patches in total). Training patches were generated using arbitrary inpainting masks with randomly varying shapes, sizes, and positions to ensure diverse lesion synthesis scenarios (see below).

**Architecture and Training Strategy**

PathoGen is based on the Latent Diffusion Model (LDM) framework [33], which synthesizes high-resolution images by operating in a compressed latent space. It combines a pre-trained Variational Autoencoder (VAE) and a denoising U-Net, denoted as $\epsilon_\theta(z_t, t)$. The VAE consists of an encoder $E$ and a decoder $D$, which map between the image and latent spaces. Given an image $x$, the latent representation is computed as $z_0 = E(x)$. During training, Gaussian noise $\epsilon \sim \mathcal{N}(0, I)$ is added to simulate the forward diffusion process, yielding a noisy latent $z_t$. The model is trained to denoise $z_t$ by minimizing the standard LDM objective:

$$\mathcal{L}_{\text{LDM}} := \mathbb{E}_{z_0=E(x),\epsilon\sim\mathcal{N}(0,I),t}[\| \epsilon - \epsilon_\theta(z_t, t) \|_2^2]$$

Through iterative denoising, the model learns to reverse the diffusion process and generate meaningful latent representations, which are then decoded to image space via $D(z_0)$.

PathoGen employs a shared VAE encoder for both benign tissue and lesion reference patches, ensuring that all inputs are embedded within a unified latent space to enable coherent integration during synthesis **(Figure 1)**. Inspired by recent advances demonstrating the effectiveness of simple concatenation-based conditioning in diffusion models, we adopt a similar strategy that leverages direct visual feature fusion without relying on additional modality-specific encoders [34]. This architectural choice not only reduces the overall parameter count—leading to improved training efficiency and memory usage—but also facilitates spatial and structural consistency across modalities. In addition, unlike text-to-image generation tasks that necessitate cross-modal semantic alignment, our task demands pixel-level spatial coherence between input tissue image and inpainted pathological structures. Therefore, we remove cross-attention modules and text encoders entirely, as textual conditioning is unnecessary for lesion inpainting where all relevant information is contained within the visual inputs.

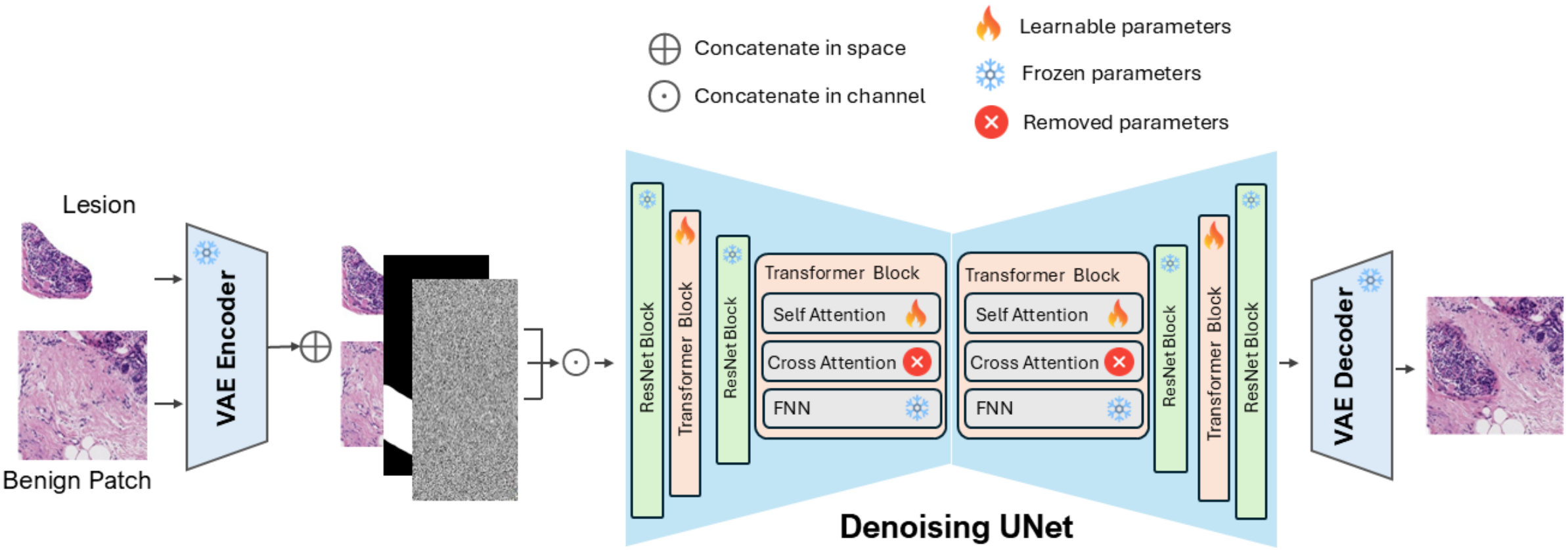

**Figure 1. Overview of the PathoGen architecture for lesion inpainting in histopathology images.** PathoGen spatially concatenates the benign tissue image with the lesion reference, ensuring they remain in the same latent feature space throughout the diffusion process. Only self-attention parameters providing global tissue-lesion interaction are learnable during training. Unnecessary cross-attention for text interaction is removed, and no additional conditions such as tissue segmentation are required. After denoising, the output latent is decoded back into image space using a frozen VAE decoder. The final output is a high-fidelity histopathology image with a synthesized lesion that is seamlessly integrated into the surrounding tissue.

To address the scarcity of paired benign-lesion datasets in histopathology, PathoGen adopts a self-supervised training strategy based on inpainting. The model learns to reconstruct randomly masked regions of real histopathology images without requiring explicit lesion annotations. For each 1024×1024 image patch $I \in \mathbb{R}^{3\times H\times W}$, a random arbitrary region is selected and expanded by a margin $\delta \sim \mathcal{U}(50,200)$ pixels in all directions to create the final inpainting mask. This gap between the original region and the visible mask boundary encourages the model to synthesize smooth, natural transitions at lesion boundaries and prevents overfitting to edge pixels, thereby avoiding trivial reconstructions. To enable dynamic control during inference, classifier-free guidance [35] is incorporated into training. With a probability of 10%, conditional inputs (masked benign image and lesion reference) are replaced with learned null embeddings. This allows the model to operate both conditionally and unconditionally, facilitating inference-time guidance scaling.

At inference, PathoGen requires only a benign histopathology image, a lesion reference patch, and a binary target mask specifying lesion placement; no preprocessing or additional conditioning is needed. The benign image is masked at the target region and encoded, together with the lesion reference, into the shared VAE latent space. The two latents and the resized mask are concatenated spatially to preserve positional alignment between the benign context and the lesion reference, then passed with initial Gaussian noise through the denoising U-Net over $T$ timesteps. The resulting latent is decoded to produce the final image with the inpainted lesion. Full equations and tensor dimensions are provided in **Supplementary Methods**. PathoGen generates one 1024×1024 patch in approximately 12 seconds on a single NVIDIA RTX 6000 GPU.

**Expert Pathologist Assessment**

We conducted a blinded reader study with six board-certified pathologists spanning diverse subspecialties and levels of experience: lymphoid pathology (15 years), breast pathology (12 years), anatomical pathology (10 years), general pathology (2 years), and two veterinary pathologists specializing in anatomic pathology (7 and 20 years). PathoGen was benchmarked against four generative baselines: a Conditional GAN (CGAN) [36], Stable Diffusion 1.5 [33], and two general-purpose commercial image-generation models, GPT-Image (OpenAI) and NanoBanana (Google). For the commercial models, we used their official cloud-based APIs, providing each with the same masked benign image, lesion reference, and target region supplied to the other methods. All learning-based baselines were trained and evaluated on identical data splits to ensure fair comparison.

The assessment comprised two parts (full protocol in **Supplementary Methods**). In Part 1 (Reality Assessment), each reviewer classified 120 images—an equal mixture of real and PathoGen-inpainted histopathology images, balanced across the four datasets—as real or AI-generated, rating their confidence on a 1–5 scale. From these responses we computed overall accuracy, sensitivity, specificity, precision, and F1 score, with chance-level discrimination defined as 50%. In Part 2 (Comparative Evaluation), each reviewer ranked the inpainted candidates from all five methods across 40 image sets by realism (color fidelity, texture, structural plausibility, and tissue blending), assigning a unique rank to each candidate. We summarized these rankings using the win rate, defined as the proportion of the 240 reader-set evaluations (6 readers × 40 sets) in which a method was ranked most realistic.

## Results

### PathoGen produces more realistic and histologically accurate lesion inpainting

To evaluate the quality and realism of lesion synthesis, we compared PathoGen against state-of-the-art generative baselines. **Figure 2A** demonstrates PathoGen's ability to generate high-fidelity lesion inpainting across four distinct histopathology datasets. For the PUMA melanoma dataset, PathoGen synthesized lesions with natural cellular density and preserved nuclear morphology characteristic of melanocytic proliferations. On the RINGS prostate dataset, the model generated glands with realistic lumen formation, appropriate epithelial layering, and natural stromal interfaces. For the TIGER breast cancer dataset, PathoGen inpainted tumor regions while maintaining complex spatial relationships between tumor cells, stromal components, and infiltrating lymphocytes, with natural nuclear pleomorphism and appropriate tissue density gradients. The KPI kidney dataset presented the most challenging synthesis task owing to its highly specialized glomerular architecture; here PathoGen generated lesions with preserved mesangial architecture, realistic capillary wall thickness, and natural Bowman's space relationships. Across all datasets, PathoGen handled diverse lesion sizes while maintaining tissue-specific staining characteristics and achieving seamless integration with surrounding tissue.

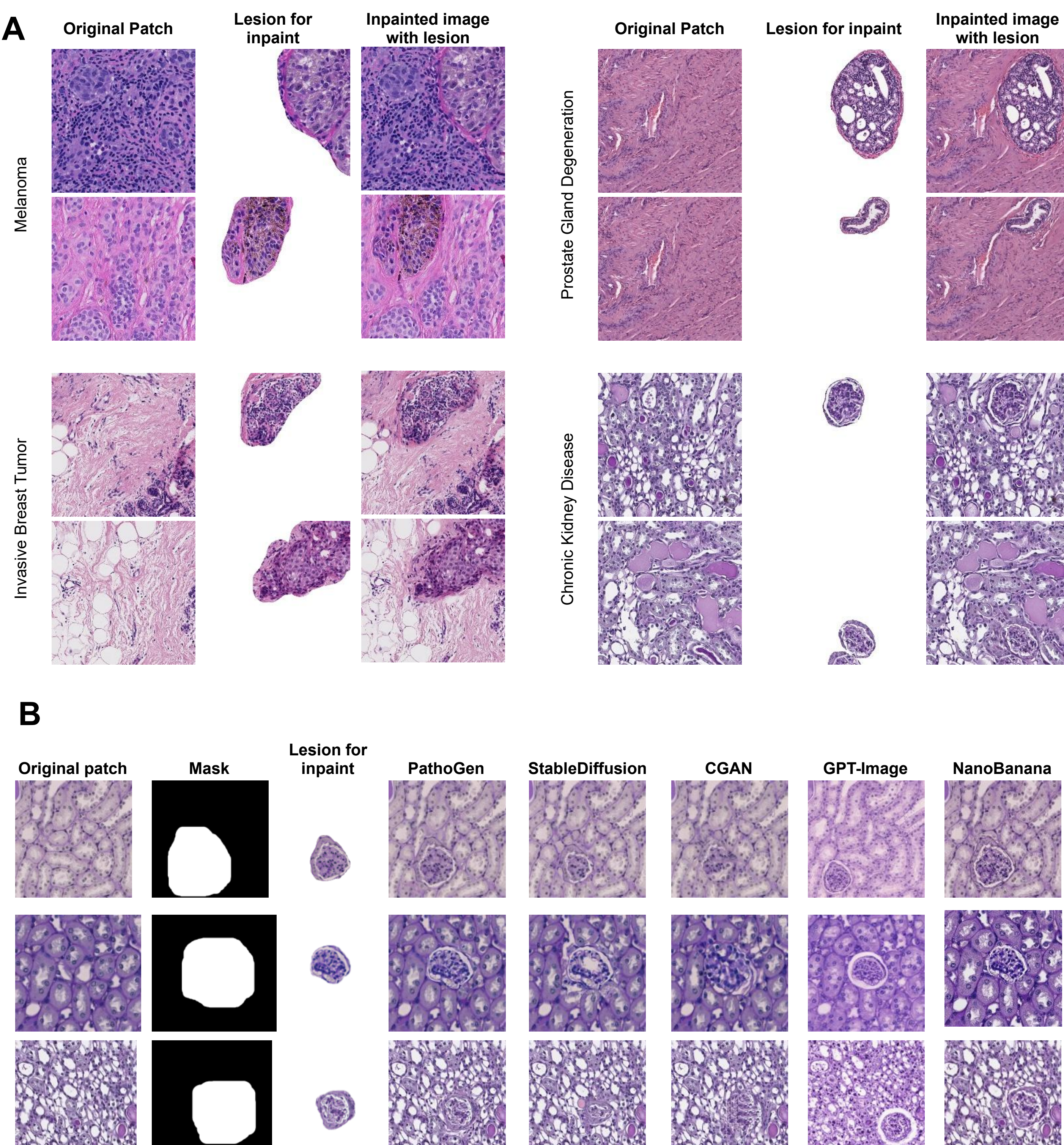

**Figure 2. PathoGen generates high-quality lesion inpainting across diverse histopathology datasets.** (A) Examples of PathoGen lesion synthesis across four tissue types (PUMA melanoma, RINGS prostate, TIGER breast, KPI kidney), showing the original benign patch, the lesion reference used for inpainting, and the resulting inpainted image. (B) Side-by-side comparison of all methods on representative kidney glomeruli, showing the original patch, binary mask, lesion reference, and outputs of PathoGen, Stable Diffusion, CGAN, GPT-Image, and NanoBanana. PathoGen produces more realistic tissue architecture and more seamless boundary integration than the baseline methods.

Side-by-side comparisons on representative kidney glomeruli examples (**Figure 2B**) revealed distinct performance differences across the five methods. PathoGen consistently generated lesions that closely matched the structural organization of the reference lesions, preserving characteristic capillary tuft architecture, appropriate mesangial matrix distribution, realistic nuclear spacing, and natural tissue density gradients. In contrast, CGAN produced noticeable structural distortions, including loss of capillary loop definition, irregular cellular distribution, and color inconsistencies, and struggled particularly at boundary regions, creating sharp discontinuities and color mismatches with surrounding tissue. Stable Diffusion showed intermediate performance, with better structural preservation than CGAN but persistent artifacts in mesangial regions, inconsistent cellular density, and subtle boundary imperfections. The general-purpose commercial models, GPT-Image and NanoBanana, generated visually plausible tissue but frequently altered global staining tone and failed to reproduce dataset-specific glomerular architecture, reflecting their lack of domain-specific training. A key advantage of PathoGen was its superior handling of tissue boundaries and contextual integration: inpainted lesions blended with the surrounding parenchyma without the abrupt discontinuities or color mismatches frequently observed in baseline outputs. This seamless integration is critical for histopathology applications, as abrupt boundaries would be readily identified as artifacts by pathologists and could confound downstream diagnostic models. Consistent with these qualitative findings, PathoGen achieved the lowest Fréchet Inception Distance (FID) [37] and Kernel Inception Distance (KID) [38] scores across all four datasets, outperforming both CGAN and Stable Diffusion and confirming that its outputs are distributionally closest to real tissue **(Supplementary Table 2)**.

**Expert pathologists achieve low accuracy in distinguishing PathoGen-synthesized lesions from real tissue**

To assess perceptual realism, six board-certified pathologists completed the two-part blinded reader study described in Methods. In Part 1 (Reality Assessment), reviewers classified a mixture of real and AI-generated images as real or synthetic. Across all reviewers, the mean discrimination accuracy was 57.75%, which, while statistically above the 50% chance level (720 trials; $p < 0.001$), reflects only low overall discrimination accuracy and indicates that synthetic lesions were difficult to distinguish from real tissue. Individual accuracies ranged narrowly from 52.6% to 61.3%, with all reviewers—regardless of subspecialty or years of experience—achieving only modest accuracy. Aggregate prediction metrics were similarly close to chance (sensitivity 54.60%, specificity 60.80%, precision 58.70%, F1 score 56.60%), confirming that performance was not driven by a systematic bias toward labeling images as real or synthetic **(Figure 3A, B)**. Self-reported confidence was weakly calibrated: classification accuracy increased with stated confidence (from chance level at confidence 1–2 to ~71% at the highest confidence level), yet nearly 30% of maximally confident judgments remained incorrect, and per-reader accuracy stayed near chance across all subspecialties and experience levels (**Supplementary Figure S1**).

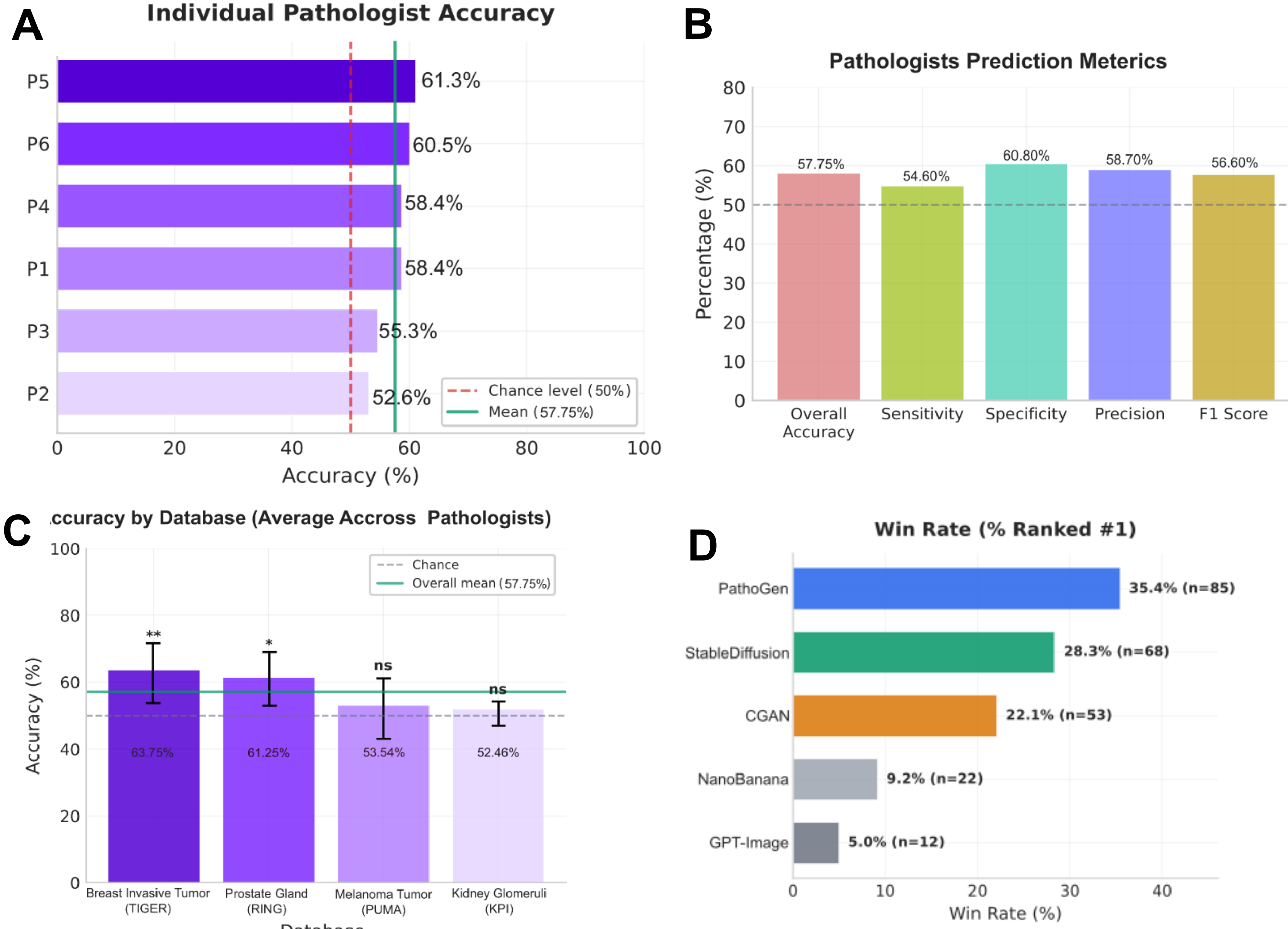

**Figure 3. Expert pathologist assessment of synthetic lesion realism.** Results of the blinded reader study with six board-certified pathologists. (Top left) Individual discrimination accuracy in Part 1, ranging from 52.6% to 61.3%, relative to the 50% chance level (dashed line) and the 57.75% group mean (solid line). (Top right) Mean accuracy by dataset, averaged across pathologists; discrimination significantly exceeded chance for breast (TIGER, 63.75%, **) and prostate (RINGS, 61.25%, *) but not for melanoma (PUMA, 53.54%, n.s.) or kidney (KPI, 52.46%, n.s.). Error bars denote 95% confidence intervals; *p < 0.05, **p < 0.01, n.s. = not significant. (Bottom left) Aggregate Part 1 prediction metrics (overall accuracy, sensitivity, specificity, precision, F1), all close to the chance level (dashed line). (Bottom right) Part 2 win rate, defined as the proportion of image sets in which each method was ranked most realistic; PathoGen achieved the highest win rate (35.4%, n = 85), followed by Stable Diffusion, CGAN, NanoBanana, and GPT-Image.

Discrimination accuracy varied by tissue type. Pathologists distinguished synthetic from real images significantly above chance for the TIGER breast (63.75%, $p < 0.01$) and RINGS prostate (61.25%, $p < 0.05$) datasets, but performed at chance for the PUMA melanoma (53.54%, n.s.) and KPI kidney (52.46%, n.s.) datasets **(Figure 3C)**. This pattern suggests that PathoGen's synthesis is most convincing for tissues with highly repetitive or specialized architecture, while subtle cues in breast and prostate morphology occasionally betrayed synthetic origin.

In Part 2 (Comparative Evaluation), reviewers ranked the inpainted candidates from all five methods for overall realism, color fidelity, texture, structural plausibility, and tissue blending. PathoGen achieved the highest win rate, being ranked most realistic in 35.4% of image sets (n

= 85), ahead of Stable Diffusion (28.3%, n = 68) and CGAN (22.1%, n = 53). The general-purpose commercial models ranked lowest, with NanoBanana and GPT-Image winning only 9.2% (n = 22) and 5.0% (n = 12) of sets, respectively **(Figure 3D)**. Together, the two parts of the reader study indicate that PathoGen-synthesized lesions are distinguished from real tissue with only low accuracy by expert reviewers and demonstrated superior performance compared to competing generative methods across multiple quality criteria.

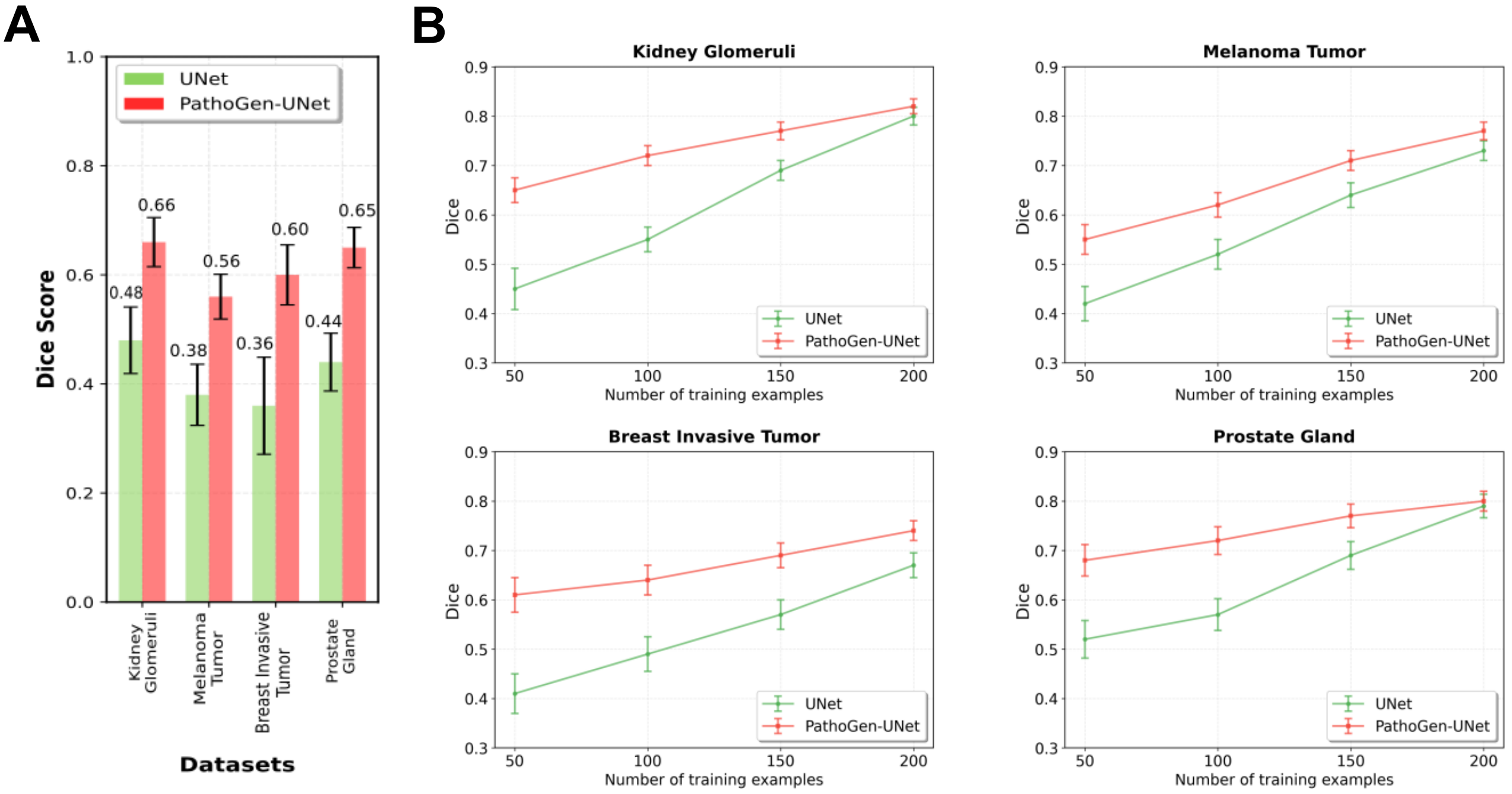

**Figure 4. Data augmentation with PathoGen-generated lesions enhances segmentation accuracy.** (A) UNet segmentation Dice scores for models trained on original data only (50 real patches; green) versus PathoGen-augmented data (50 real + 150 synthetic patches; red), evaluated on 50 held-out real patches per dataset. (B) Learning curves plotting Dice score against the number of real training samples (50, 100, 150, 200) for baseline UNet (green) and PathoGen-augmented UNet (red), across all four tissue types. In every condition, models were evaluated on 50 separate unseen real patches. Error bars denote standard deviation. PathoGen augmentation provides consistent improvements, with the largest gains at the smallest sample sizes.

### Synthetic lesions from PathoGen improve segmentation performance

To assess practical value, we trained UNet segmentation models under controlled data conditions and measured the Dice score on a fixed set of 50 unseen real patches per dataset. Three experiments were conducted. First, we compared a baseline trained on 50 real patches against a PathoGen-augmented model trained on 50 real patches plus 150 PathoGen-synthesized patches (200 patches total). Because each synthesized lesion is defined by a known mask, PathoGen automatically supplied pixel-level ground truth for all 150 synthetic patches, requiring no additional manual annotation. Second, we constructed learning curves by varying the amount of real training data (50, 100, 150, and 200 patches) with and without PathoGen augmentation, evaluating on the same 50 held-out real patches in every condition. Third, we benchmarked PathoGen against alternative augmentation strategies under matched conditions; this comparison is reported in the following section.

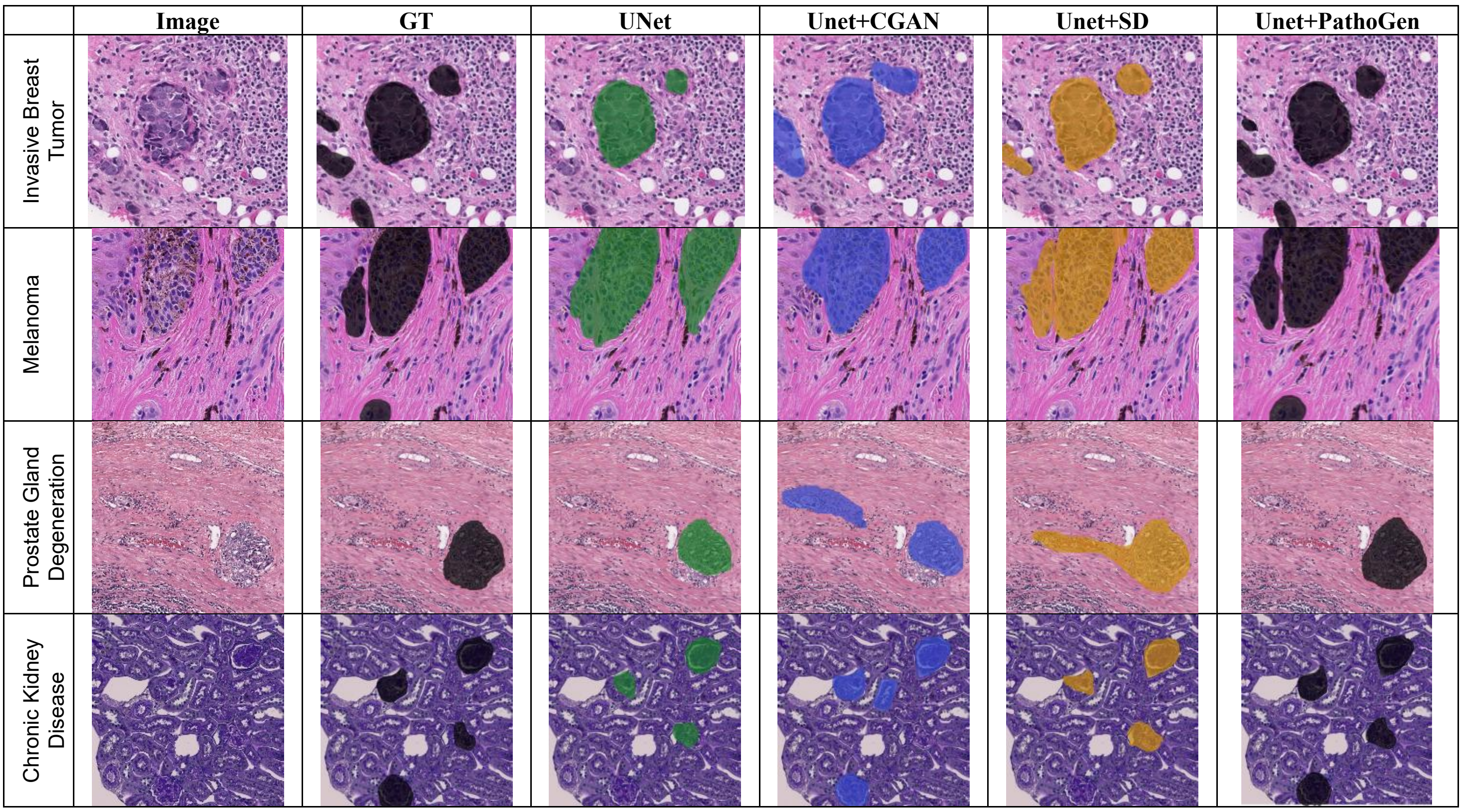

**Figure 5. Qualitative segmentation results on representative test patches.** For each dataset (TIGER, PUMA, RINGS, KPI), columns show the input image, ground-truth mask (GT), and predicted segmentations from the baseline UNet and from UNet trained with CGAN-, Stable Diffusion- (SD), and PathoGen-augmented data. PathoGen-augmented models more closely match the ground truth and produce fewer false-positive regions than the baseline and competing augmentation methods.

In the first experiment, models trained with PathoGen-augmented data showed substantial improvements across all four tissue types **(Figure 4A)**. Kidney glomeruli segmentation improved from a Dice score of 0.48 to 0.66, melanoma tumor from 0.38 to 0.56, breast invasive tumor from 0.36 to 0.60, and prostate gland from 0.44 to 0.65. The consistent improvement across datasets indicates that PathoGen generates lesions that are not only visually realistic but also carry morphological information useful for downstream tasks. The learning-curve experiment **(Figure 4B)** shows that PathoGen-augmented models consistently outperform baseline models across all data regimes and tissue types. The performance gap was most pronounced at small sample sizes and gradually narrowed as more real data became available, confirming that synthetic augmentation provides maximum value in the low-data regime where additional annotation is most costly. The magnitude and persistence of the gains varied by tissue: breast invasive tumor maintained a substantial gap even at the largest sample size, suggesting that PathoGen supplies complementary morphological diversity beyond the original dataset, whereas the advantage for kidney glomeruli and prostate gland diminished with increasing real data. Representative qualitative segmentation results are shown in **Figure 5**, where PathoGen-augmented models more accurately delineate target structures and produce fewer false-positive regions than the baseline UNet or models augmented with CGAN or Stable Diffusion.

### Morphological augmentation outperforms geometric transformations

To isolate the contribution of PathoGen's morphological synthesis, we benchmarked it against traditional geometric transformations (rotation, flipping, translation, and their combination) and against the generative baselines CGAN and Stable Diffusion. All strategies augmented a common base of 50 real patches with 150 synthetic samples before training UNet segmentation models. As shown in **Table 1**, PathoGen achieved the highest Dice score on all four tissue types (0.66 kidney, 0.56 melanoma, 0.60 breast, 0.65 prostate), outperforming both geometric and generative baselines. Among geometric transformations, rotation produced the most noticeable gains, with flipping and translation contributing smaller increments and their combination yielding the best geometric result; however, all remained well below PathoGen. Generative baselines improved on geometric methods—Stable Diffusion (0.52–0.63) exceeded CGAN (0.51–0.61)—but still trailed PathoGen across every dataset. The limited effectiveness of geometric augmentations reflects their inability to generate novel pathological variation, while the gap relative to PathoGen among the generative methods likely stems from weaker spatial control and insufficient awareness of tissue context during synthesis.

**Table 1. Segmentation performance comparison across augmentation methods.** Dice scores (mean ± s.d.) for UNet models trained with different augmentation approaches on four tissue types. Vanilla denotes the baseline trained on 50 real patches with no augmentation. Geometric methods (Rotate, Flip, Translate, Combine) expand this base to 200 patches using the corresponding transformations, while the generative methods (CGAN, Stable Diffusion, PathoGen) add 150 synthetic samples to the same 50 real patches (200 patches total). PathoGen consistently outperforms both traditional geometric augmentations and the generative baselines across all four tissue types.

| Augmentation | Kidney Glomeruli | Melanoma Tumor | Breast Invasive Tumor | Prostate Gland |
|---|---|---|---|---|
| **Vanilla** | 0.48 ± 0.021 | 0.38 ± 0.020 | 0.36 ± 0.038 | 0.44 ± 0.018 |
| **Rotate** | 0.56 ± 0.012 | 0.51 ± 0.018 | 0.52 ± 0.015 | 0.57 ± 0.015 |
| **Flip** | 0.54 ± 0.018 | 0.49 ± 0.022 | 0.49 ± 0.017 | 0.57 ± 0.017 |
| **Translate** | 0.55 ± 0.014 | 0.50 ± 0.019 | 0.48 ± 0.016 | 0.54 ± 0.016 |
| **Combine** | 0.58 ± 0.016 | 0.52 ± 0.011 | 0.53 ± 0.019 | 0.59 ± 0.019 |
| **CGAN** | 0.60 ± 0.013 | 0.51 ± 0.017 | 0.54 ± 0.014 | 0.61 ± 0.014 |
| **StableDiffusion** | 0.62 ± 0.011 | 0.52 ± 0.020 | 0.55 ± 0.018 | 0.63 ± 0.018 |
| **PathoGen** | **0.66 ± 0.019** | **0.56 ± 0.023** | **0.60 ± 0.021** | **0.65 ± 0.021** |

## Discussion

### PathoGen addresses the dual bottleneck of data scarcity and annotation cost in computational pathology

This study introduces PathoGen, a diffusion-based generative model that addresses a fundamental challenge in computational pathology: the scarcity of expert-annotated training data for AI model development. Our results demonstrate that PathoGen generates high-fidelity lesion inpainting across diverse tissue types while maintaining realistic cellular architectures, natural tissue boundaries, and authentic staining characteristics. When used for data augmentation, PathoGen-generated synthetic lesions consistently improve downstream segmentation performance across four clinically relevant histopathology datasets, with the most substantial gains observed in low-data regimes where acquiring additional annotations is most challenging [39].

A key practical advantage of PathoGen is that since inpainted lesion regions are defined by known masks, the model automatically provides pixel-level annotations for each generated patch. This capability enables the generation of large numbers of diverse, fully-annotated training samples without the need for manual expert annotation—a process that typically requires hours of pathologist time per image [40]. By simultaneously generating both synthetic images and their corresponding ground-truth annotations, PathoGen addresses the dual challenge of data scarcity and annotation cost that constrains AI development in computational pathology [41].

The superiority of PathoGen over existing approaches stems from several key advantages inherent to diffusion-based architectures. Unlike GANs, which suffer from training instability and mode collapse [42], the iterative refinement process of diffusion models enables more stable training and broader coverage of the morphological feature space [20]. This translates to greater diversity in generated lesion morphologies and more consistent preservation of fine-grained histological details. Furthermore, PathoGen's conditioning mechanism provides precise spatial control over lesion placement while ensuring contextual awareness of surrounding tissue—a critical requirement for generating lesions that seamlessly integrate with benign parenchyma. Our comparative analysis reveals that while general-purpose diffusion models like Stable Diffusion demonstrate reasonable performance, PathoGen's domain-specific architecture and training strategy yield substantially better results, highlighting the importance of tailoring generative models to the unique requirements of histopathology [43].

### Expert pathologists cannot reliably distinguish PathoGen-synthesized lesions from real tissue

Beyond quantitative image-quality metrics, the blinded reader study with six board-certified pathologists provides direct evidence that PathoGen's synthetic lesions are perceptually realistic. In the reality-assessment task, mean discrimination accuracy was only 57.75%, marginally above the 50% chance level, and no individual reviewer—regardless of subspecialty or years of experience—approached reliable discrimination. The balanced aggregate metrics (sensitivity 54.60%, specificity 60.80%, precision 58.70%, F1 56.60%) further indicate that this near-chance performance was not an artifact of a systematic bias toward labeling images as real or synthetic, but rather reflects genuine difficulty in telling synthetic lesions apart from authentic tissue. In the comparative-evaluation task, PathoGen achieved the highest win rate (35.4%), ahead of Stable Diffusion (28.3%) and CGAN (22.1%), while the general-purpose commercial models GPT-Image (5.0%) and NanoBanana (9.2%) ranked lowest. The poor showing of these powerful but domain-agnostic models reinforces our central finding: realistic histopathological synthesis requires domain-specific training and tissue-aware conditioning rather than general visual plausibility alone. Together, the two parts of the reader study show that PathoGen is not only difficult to distinguish from real tissue but is also preferred over competing generative methods by expert reviewers, providing the kind of human-expert validation that quantitative metrics alone cannot supply.

Notably, discrimination accuracy varied by tissue type. Pathologists distinguished synthetic from real images above chance for the TIGER breast (63.75%) and RINGS prostate (61.25%) datasets, but performed at chance for the PUMA melanoma (53.54%) and KPI kidney (52.46%) datasets. This pattern suggests that PathoGen's synthesis is most convincing for tissues with highly repetitive or specialized architecture, whereas the more heterogeneous spatial relationships of breast and prostate morphology occasionally leave subtle cues that betray synthetic origin. Identifying and correcting these residual, tissue-specific artifacts represents a concrete direction for further improving generation fidelity.

**Morphological augmentation captures tissue-specific constraints that geometric transformations cannot represent**

Our comparative analysis against traditional geometric transformations reveals fundamental limitations of conventional augmentation approaches in histopathology [12]. While rotation, flipping, and translation provide modest performance gains, they cannot generate novel morphological variations or bridge gaps in the training distribution. This limitation is particularly acute in histopathology where diagnostic features span multiple spatial scales and subtle morphological variations carry critical clinical significance. PathoGen, in contrast, synthesizes entirely new lesion instances with realistic cellular arrangements, appropriate nuclear characteristics, and natural tissue architecture—generating morphological diversity that extends beyond simple recombinations of existing visual information.

The superiority of morphological augmentation over geometric transformations also reflects the inherent properties of histopathological images. Unlike natural images where objects can appear at arbitrary orientations and scales, tissue architecture exhibits strong spatial priors and anatomical constraints [44]. Glomerular capillary loops, glandular lumens, and tumor-stromal interfaces follow characteristic organizational patterns that are not fully captured by simple geometric transformations. PathoGen's learned generative process respects these tissue-specific constraints while introducing biologically plausible variations, resulting in synthetic samples that better represent the true distribution of pathological presentations. The fact that PathoGen-augmented models also produced fewer false-positive regions than models augmented with CGAN or Stable Diffusion (Figure 5) indicates that this added morphological diversity translates into more discriminative, rather than merely more numerous, training examples.

### Computational efficiency and expert validation remain key challenges for clinical translation

Despite promising results, two key limitations should be addressed in future work. First, computational requirements for diffusion-based generation exceed those of traditional augmentation approaches. PathoGen requires approximately 12 seconds per 1024×1024 patch on a modern NVIDIA RTX6000 GPU, compared to milliseconds for geometric transformations. While this overhead is negligible compared to expert annotation time (which typically requires hours per whole-slide image), it may constrain real-time augmentation during training or deployment in computational resource-limited settings. Future work exploring distillation techniques [45] or accelerated sampling methods [45] could mitigate this limitation while preserving generation quality.

Second, although our reader study establishes that PathoGen's lesions are perceptually realistic and difficult to distinguish from real tissue, perceptual realism is not equivalent to diagnostic fidelity [46]. Our evaluation assessed overall realism, structural plausibility, and tissue blending, but did not test whether synthetic lesions faithfully reproduce clinically defined attributes such as tumor grade, subtype-specific features, or the presence of diagnostically misleading artifacts. The tissue-specific discrimination differences noted above further suggest that residual artifacts persist for certain morphologies. Future studies should therefore extend expert evaluation beyond realism to grading- and subtype-level correctness, ideally coupling reader assessment with quantitative diagnostic concordance, to ensure that synthetic data does not introduce systematic biases into downstream diagnostic models. Such validation will be essential before generative augmentation can be safely incorporated into clinically deployed pathology AI pipelines.

## Conclusion

PathoGen demonstrates that diffusion-based generative models, when fine-tuned for domain-specific histopathology images, can effectively address data scarcity challenges through controllable, high-fidelity image synthesis. By generating realistic lesions that preserve complex tissue architectures and seamlessly integrate with surrounding parenchyma, PathoGen enables substantial improvements in downstream segmentation tasks, particularly in low-data regimes where manual annotation is most challenging. Critically, a blinded reader study with six board-certified pathologists confirmed that PathoGen's synthetic lesions are difficult to distinguish from real tissue—achieving near-chance discrimination accuracy—and are preferred over competing generative methods, providing direct expert validation of their perceptual realism. The consistent performance across diverse tissue types and superiority over both traditional augmentation methods and alternative generative approaches establish PathoGen as a promising tool for enhancing AI model development in histopathology. As computational pathology continues advancing toward clinical deployment, methods like

PathoGen that can efficiently generate high-quality, expert-validated training data while respecting the unique morphological complexity of tissue images will play an increasingly important role in developing robust, generalizable diagnostic AI systems.

**Conflict of interest**

The authors of this study declare that they do not have any conflict of interest.

**Data availability statement**

The data we used are publicly available and have already been referenced in our manuscript. The training and inference code is available at https://github.com/ClinicalAI/PathoGen. Additionally, the fine-tuned model weights are available on our Hugging Face page: https://huggingface.co/mkoohim/PathoGen.

**Funding**

This work was conducted in the JC STEM Lab of Innovative Medical Imaging Research funded by The Hong Kong Jockey Club Charities Trust (No specific grant number).

**Author Contributions Statement**

MK, MNM, and KTB conceived the idea. MK and MNM prepared the training and test datasets and implemented the diffusion model. MK and MNM evaluated the model. RAU, OU, MM, PI, FL, and AF conducted the blinded pathologist reader study, reviewing and ranking the histopathology images according to the study protocol. MK wrote the manuscript, and MNM and KTB provided comments and revisions. KTB provided funding for the project.